\documentclass{article}

\usepackage{microtype}
\usepackage{graphicx}
\usepackage{subcaption}
\usepackage{booktabs}

\PassOptionsToPackage{hyphens}{url}\usepackage{hyperref}

\usepackage[preprint]{icml2026}

\usepackage{amsmath}
\usepackage{amssymb}
\usepackage{mathtools}
\usepackage{amsthm}

\usepackage[capitalize,noabbrev]{cleveref}

\theoremstyle{plain}

\theoremstyle{definition}

\theoremstyle{remark}

\usepackage[textsize=tiny]{todonotes}

\usepackage[dvipsnames]{xcolor}
\usepackage{fontawesome}
\usepackage[frozencache]{minted}
\usepackage{bm}
\usepackage[most]{tcolorbox}
\usepackage{setspace}
\usepackage{placeins}
\usepackage{wrapfig}
\usepackage{multirow}

\usepackage[T1]{fontenc}

\icmltitlerunning{Unified Zero-Shot Time Series Forecasting: A Darts Foundation}

\graphicspath{{figures/}}

\definecolor{darts}{RGB}{0, 0, 245}
\DeclareRobustCommand{\Logo}{%
  \begingroup\normalfont
  \vspace{-0.2em}%
  \raisebox{-0.3em}{%
  \includegraphics[height=1.2em]{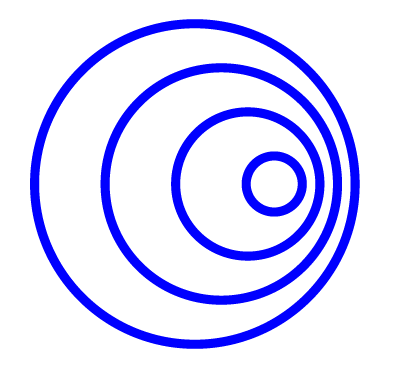}%
  }%
  \kern 0.3em%
  \endgroup
}
\newcommand*{\Darts}{\textbf{Dar\textcolor{darts}{ts}}}%
\newcommand{\github}{\raisebox{0pt}{\faGithub}}%
\newcommand{\book}{\raisebox{0pt}{\faBook}}%
\definecolor{darkblue}{HTML}{0077b6}
\newcommand{\gh}[1]{}

\usepackage{enumitem}

\tcbset{fontlower=\tiny}
\newenvironment{quickstart}[1][]
{\begin{tcolorbox}[colback=darts!5,colframe=darts!80,title=#1]}
{\end{tcolorbox}}

\begin{document}

\twocolumn[
  \icmltitle{Unified Zero-Shot Time Series Forecasting: \\
  A \Darts\Logo Foundation}

  \icmlsetsymbol{equal}{*}

  \begin{icmlauthorlist}
    \icmlauthor{Zhihao Dai}{equal,oxford}
    \icmlauthor{Dennis Bader}{equal,unit8}
    \icmlauthor{Alain Gysi}{unit8}
    \\{$^*$}{Equal contribution}
  \end{icmlauthorlist}

  \icmlaffiliation{oxford}{Energy and Power Group, Department of Engineering Science, University of Oxford, UK}
  \icmlaffiliation{unit8}{Unit8 SA, Switzerland}

  \icmlcorrespondingauthor{Dennis Bader}{dennis.bader@unit8.com}
  
  {\centerline{\github\,\,\url{https://github.com/unit8co/darts}}}

  \icmlkeywords{Machine Learning, open-source tools, time series forecasting, foundation models}

  \vskip 0.3in
]

\printAffiliationsAndNotice{}

\begin{abstract}
  Since its initial release in 2020, \Darts~has become a widely used open-source Python library for time series analysis.
  A series of foundation models have recently claimed accuracy improvements in zero-shot forecasting, promising a paradigm shift from training custom models to harnessing pre-trained general-purpose forecasters.
  Foundation models, however, are often released as isolated packages with fragmented interfaces and limited interoperability with common tooling, making joint evaluation and integration within complete pipelines difficult.
  In \Darts, we developed a unified \texttt{FoundationModel} class collection (Chronos-2, TimesFM 2.5, TiRex, PatchTST-FM) that provides standardized, full-cycle forecasting interfaces with minimal external dependencies for integrating foundation models into the ecosystem.
  Existing \Darts~pipelines can now use foundation models with only a name change; new pipelines can use them for zero-shot or fine-tuned forecasting, uncertainty estimation, and backtesting, combined with data processing and evaluation tooling, all within a unified framework.
\end{abstract}

\begin{figure}[H]
\vspace{-12pt}
\begin{quickstart}[Quickstart]
\begin{minted}{python}
# Install via `pip install darts[torch]`
from darts.datasets import WineDataset
from darts.models import TimesFM2p5Model

series = WineDataset().load().astype("f")
model = TimesFM2p5Model(
    input_chunk_length=36,
    output_chunk_length=12
).fit(series)
forecast = model.predict(n=12, series=series)
\end{minted}
\end{quickstart}
\end{figure}

\section{Introduction}
Pre-trained on massive real-world and synthetic data~\cite{godahewa2021monash,aksu2024gifteval,goswami2024moment}, time series foundation models (TSFMs)~\cite{garza2023timegpt1,goswami2024moment,woo2024moirai,ansari2024chronos,das2024timesfm,auer2025tirex,ansari2025chronos2,graf2025flowstate,wen2026patchtstfm,fu2026reverso} capture diverse temporal patterns and are emerging as general-purpose forecasters applicable to unseen time series without further training, i.e., zero-shot forecasting.
All of them have successively claimed improvements over series- and task-specific models on public benchmarks~\cite{wu2023timesnet,aksu2024gifteval,shchur2025fev} and envisioned a future where TSFMs replace specialized models for most, if not all, forecasting tasks.
Successful applications include forecasting for respiratory disease incidence~\cite{ecdc2026respicast}, retail sales~\cite{bruned2025showdown,pokrzywa2026enhancing}, and defect detection~\cite{zhang2025defect}.

Yet that vision will remain a vision should TSFMs continue to live in isolated packages with neither unified APIs nor interoperability with common time series tooling.
Existing forecasting pipelines desiring TSFMs must be retrofitted for idiosyncratic interfaces of each TSFM.
Dependency conflicts and divergent behaviors (e.g., probabilistic sampling) often ensue with multiple TSFMs.
Further, new pipelines using TSFMs would still demand common processing (e.g., resampling) and evaluation (e.g., metrics) tooling, which is either unavailable or has been rewritten in TSFM packages, leading to fragmentation and duplicated efforts.

Since its initial release in 2020, \Darts~\cite{herzen2022darts} has aspired to be a user-friendly, versatile machine learning library for time series data, providing unified APIs for forecasting, classification, and anomaly detection and encompassing data transformation and evaluation tooling.
While TSFMs are still in their infancy and face concerns about generalizability~\cite{mulayim2024tsfm} and benchmarking practices~\cite{bergmeir2024llms}, we believe in the value of a common TSFM foundation facilitating user adoption and fair comparison, and that \Darts~is well-positioned to provide it.

\setlength{\tabcolsep}{5pt}
\renewcommand{\thefootnote}{\fnsymbol{footnote}}
\begin{table}[t]
  \caption{Foundation model support in open-source Python libraries, including model availability, forecasting capabilities (covariates, probabilistic, and multiple series), fine-tuning capability, and explainability, as of June 2026.}
  \label{tab:comparison}
  \begin{center}
    \begin{small}
\begin{tabular}{@{}l|cccccc@{}}
\toprule
\multirow{2}{*}{Library} & \multirow{2}{*}{Models\footnotemark[1]} & \multicolumn{3}{c}{Forecasting} & \multirow{2}{*}{\begin{tabular}[c]{@{}c@{}}Fine-\\ Tuning\end{tabular}} & \multirow{2}{*}{XAI} \\ \cmidrule(lr){3-5}
 &  & Cov. & Prob. & Multi. &  &  \\ \midrule
\Darts & 4 & $\checkmark$ & $\checkmark$ & $\checkmark$ & $\checkmark$ & $\checkmark$ \\
TimeCopilot & 11 & $\times$ & $\checkmark$ & $\checkmark$ & $\checkmark$ & $\times$ \\
nixtla & 1 & $\checkmark$ & $\checkmark$ & $\checkmark$ & $\checkmark$ & $\checkmark$ \\
sktime & 11 & $\checkmark$ & $\checkmark$ & $\checkmark$ & $\checkmark$ & $\times$ \\
AutoGluon & 2 & $\checkmark$ & $\checkmark$ & $\checkmark$ & $\checkmark$ & $\times$ \\
skforecast & 4 & $\checkmark$ & $\checkmark$ & $\checkmark$ & $\times$ & $\times$ \\
Transformers & 1 & $\times$ & $\checkmark$ & $\checkmark$ & $\checkmark$ & $\times$ \\
aeon & 0 & - & - & - & - & - \\
NF & 0 & - & - & - & - & - \\
PTF & 0 & - & - & - & - & - \\ \bottomrule
\end{tabular}
    \end{small}
  \end{center}
\begin{small}
    \footnotesize
    \footnotemark[1] Models from the same family are counted as one.
\end{small}
\vspace{-8mm}
\end{table}
\renewcommand{\thefootnote}{\arabic{footnote}}

In \Darts, we developed a \texttt{FoundationModel} class collection with unified APIs and minimal external dependencies, initially comprising Chronos-2~\cite{ansari2025chronos2}\gh{2944}, TimesFM 2.5~\cite{das2024timesfm}\gh{2980}, TiRex~\cite{auer2025tirex}\gh{3038}, and PatchTST-FM~\cite{wen2026patchtstfm}\gh{3120}.
A three-liner Quickstart is shown on the cover, and \texttt{darts[torch]} is the only prerequisite installation.

Within the \Darts~ecosystem, end users can now co-deploy TSFMs with specialized models and common tooling, ensuring consistent forecasting protocols and minimizing dependency conflicts.
Under unified APIs, switching between models is as easy as a name change, and joint evaluation or ensembling becomes straightforward.
On the other end, TSFM providers can ship their models in \Darts~to reach a wide user base and reduce the effort of reinventing common tooling, knowing that we offer a level playing field for all TSFMs and a short path to end users.

\Darts~is not the only framework providing zero-shot forecasting; Table~\ref{tab:comparison} lists contemporary open-source libraries and their TSFM support to the best of our knowledge.
This includes nixtla~\cite{garza2023timegpt1}, sktime~\cite{loning2019sktime}, AutoGluon~\cite{shchur2023autogluon}, skforecast~\cite{rodrigo2026skforecast}, aeon~\cite{middlehurst2024aeon}, NeuralForecast~\cite{olivares2022neuralforecast}, PyTorch Forecasting~\cite{beitner2020pytorchforecasting}, and Transformers~\cite{wolf2020transformers}.
A notable mention is TimeCopilot~\cite{garza2025timecopilot}, an agentic toolkit that uses Language Models to orchestrate a forecasting workflow.

We do not claim that \Darts~is superior or more complete than other libraries, since each appeals to its target users in its own way.
Our contributions in this paper are:
\setlist{nolistsep}
\begin{itemize}[noitemsep]
    \item A unified \texttt{FoundationModel} API integrating four TSFMs with minimal external dependencies.
    \item Native support for covariates, probabilistic forecasting, and fine-tuning under a single interface.
    \item Model-agnostic explainability via SHAP values for all PyTorch-based models.
    \item Full interoperability with existing \Darts~tooling for evaluation, backtesting, and data processing.
\end{itemize}

\section{Design Principles and Key Features}
\subsection{\texttt{FoundationModel} Unified APIs}
The new \texttt{FoundationModel} class minimally extends \texttt{TorchForecastingModel} (no extra methods) and is the base class for all PyTorch-native TSFMs\gh{2944}.
Like all \Darts~models, it provides standardized \texttt{fit(series: TimeSeries)} and \texttt{predict(n: int)} methods and can consume and produce \texttt{TimeSeries} objects, ensuring framework-wide interoperability.
\texttt{TimeSeries} is an immutable container~\cite{herzen2022darts} housing 3D time-indexed data, \texttt{(time, components, samples)}, alongside optional fields such as metadata.
It can be converted to and from common formats such as pandas DataFrames~\cite{pandas2020pandas} and NumPy arrays~\cite{harris2020numpy}.
As shown in the Quickstart, \texttt{fit} must be called before \texttt{predict}, even for training-free zero-shot forecasting, to enforce a uniform workflow and allow for model creation and sanity checks.

Under the hood, \texttt{FoundationModel} uses the PyTorch Lightning backend~\cite{falcon2019pytorchlightning} to dispatch training and inference loops.
We expose Lightning's \texttt{Trainer} keyword arguments to users for fine-grained control, such as early stopping and hardware acceleration.
All \texttt{TorchForecastingModel} functionalities are supported, including optimized historical forecasting, error backtesting, residual computation, and model loading and saving, providing consistent behavior across models.

For TSFM providers, implementing a model requires only two steps:
(1) subclass a \texttt{PLForecastingModule} and define its \texttt{forward} pass, and (2) subclass \texttt{FoundationModel} and implement the \texttt{\_create\_model} method to return a module instance.
The rest is handled by \Darts, from data ingestion to training and inference.
For providers' convenience, we include a \texttt{HuggingFaceConnector} class to download and load safetensors-formatted~\cite{huggingface2022safetensors} models from Hugging Face Hub~\cite{huggingface2021huggingfacehub}, which has become the go-to TSFM distribution platform, but its usage is not compulsory.

\subsection{Forecasting with Covariates}
Covariates, i.e., exogenous variables, contextualize the target series to forecast and are crucial for many tasks.
Thanks to \Darts' native covariate support, covariate-aware TSFMs such as Chronos-2 can forecast with covariates out of the box.

We define \textbf{past covariates} as exogenous series observed up to the forecast start time, and \textbf{future covariates} as those known up to the forecast end time~\cite{herzen2022darts}, each contained in a \texttt{TimeSeries} instance.
Predictions can be made by simply calling \texttt{predict(n, series, past\_covariates, future\_covariates)}.
Alignment between the target series and covariates is automatically handled by \Darts~and prevents data leakage.

\subsection{Forecasting Multiple Series}
\label{sec:multi-series}
It is common to forecast multiple series simultaneously, such as product sales across multiple stores.
\Darts~allows calling \texttt{predict} on a sequence of \texttt{TimeSeries} instances, e.g., \texttt{predict(n, [series1, series2, ...])}, and returns a sequence of \texttt{TimeSeries} forecasts.
Under the hood, input series are automatically grouped into mini-batches and processed in parallel for efficiency before outputs are reassembled into the original order.

\subsection{Optimized Historical Forecasting}
Historical forecasting simulates predictions at time points in the past and is essential for robust, realistic model evaluation.
\Darts~provides a highly configurable \texttt{historical\_forecast(series, retrain: bool = True)} method, which by default retrains the model and predicts at each time point for model freshness.
When \texttt{retrain} is \texttt{False}, it simulates forecasting without retraining and groups historical time points into batches for parallel processing, which drastically improves speed.
Like in Section~\ref{sec:multi-series}, the method accepts multi-series input and returns multi-series output.
Additionally, \Darts~provides \texttt{backtest} and \texttt{residuals} methods built on top for computing historical error metrics and forecast residuals.

\subsection{Probabilistic Forecasting}
\begin{figure}[t!]
  \begin{center}
    \centerline{\includegraphics[width=\linewidth]{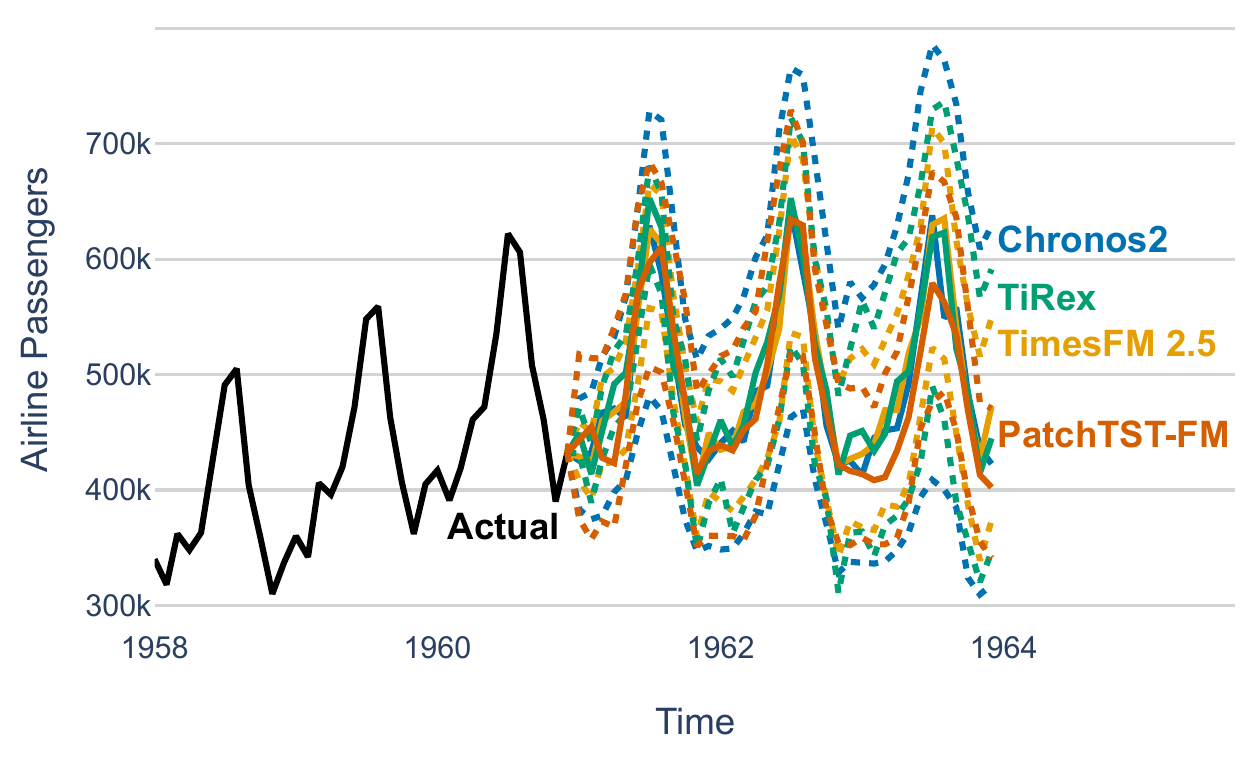}}
    \caption{Probabilistic forecasts on the air passengers dataset~\cite{box1976time}. Uncertainty intervals are shown between dotted lines, and the median is shown as solid lines. Adjusted for visualization and not indicative of actual performance.}
    \label{fig:examples}
  \end{center}
  \vspace{-10mm}
\end{figure}

Like all \texttt{TorchForecastingModel}s in \Darts, \texttt{FoundationModel} supports probabilistic forecasting, which produces uncertainty estimates rather than deterministic point forecasts and is valuable for risk-aware applications.
To enable it, a TSFM must be initialized with its likelihood model used for pre-training, which defines the forecast distribution family and its loss function.
\Darts~supports two modes of probabilistic forecasting:
\setlist{nolistsep}
\begin{enumerate}[noitemsep]
    \item \textbf{Monte Carlo Sampling}, which generates multiple forecast sample trajectories from specified distributions
    \item \textbf{Direct Parameter Prediction}, which outputs time-varying distribution parameters directly
\end{enumerate}
Figure~\ref{fig:examples} shows a probabilistic forecast example on the classic air passengers dataset~\cite{box1976time} using Chronos-2, TiRex, TimesFM 2.5, and PatchTST-FM, all of which use \texttt{QuantileRegression} likelihood.

For TSFM providers, \texttt{FoundationModel} can pair with arbitrary likelihoods, either from \Darts~or custom-defined, and samples are drawn accordingly.
While most TSFMs use quantile regression for pre-training, TSFMs using other likelihoods~\cite{cohen2025toto} can also be supported via model-likelihood pairing, enabling maximum flexibility.

\subsection{Fine-tuning for Tasks}
Beyond zero-shot forecasting, a core strength of TSFMs is their adaptability to specific tasks via fine-tuning~\cite{goswami2024moment,ekambaram2024tiny,fu2025financial}, which can boost accuracy.
\Darts~adds fine-tuning support to \texttt{FoundationModel} via the \texttt{enable\_finetuning: bool | dict[str, list[str]] | None} argument at model initialization\gh{3003}.
The argument value determines the model behavior during a \texttt{fit} call:
\setlist{nolistsep}
\begin{itemize}[noitemsep]
    \item \texttt{False} or \texttt{None} (default): model is frozen for zero-shot forecasting, and \texttt{fit} does not update parameters.
    \item \texttt{True}: all model parameters are unfrozen for deep fine-tuning, and \texttt{fit} performs fine-tuning.
    \item \texttt{dict[str, list[str]]}: only specified model parameters are frozen or unfrozen for fine-tuning, e.g., \texttt{\{"unfreeze": ["layer8.*"]\}} to only update "layer8", or \texttt{\{"freeze": ["*.attention.*"]\}} to freeze "attention" weights and fine-tune the rest. Parameter names are matched by Unix shell-style wildcards~\cite{psf2026fnmatch}.
\end{itemize}
The design covers common TSFM usage, from zero-shot to partial and full fine-tuning, and prevents model updates during a \texttt{predict} call.
Currently, parameter-efficient fine-tuning methods~\cite{hu2022lora} are not supported.

This transfer learning paradigm also benefits deep learning models in \Darts~beyond TSFMs.
For instance, one might train a base TFT model~\cite{lim2021tft} for national energy demand forecasting and fine-tune it for regional forecasting.
We therefore uplifted the \texttt{enable\_finetuning} logic to \texttt{TorchForecastingModel}, so that all PyTorch models can be fine-tuned from a loaded checkpoint.
For non-TSFMs, \texttt{None} (default) is equivalent to \texttt{True}, i.e., fully trainable by default.

\subsection{Model Explainability}
Explainability is desirable for building trust and understanding of complex models like TSFMs.
\Darts~offers \texttt{ShapExplainer}\gh{3049} for explaining forecasts of PyTorch models via SHAP values~\cite{lundberg2017shap}.
A SHAP value quantifies each input's contribution to moving a forecast away from an average forecast (i.e., the base value); its sign and magnitude indicate the direction and strength of the contribution.
By default, \texttt{ShapExplainer} uses model-agnostic Permutation SHAP, which permutes inputs to estimate SHAP values.
Deep SHAP~\cite{shrikumar2017deeplift} and Gradient SHAP~\cite{sundararajan2017gradient} were not adopted due to architectural restrictions and the lack of base values required for common visualizations.

\subsection{More Capabilities}
Beyond the aforementioned key features and common tooling such as data transformation and evaluation, \Darts~also ships many capabilities that directly benefit TSFM usage.
This includes auto-regressive forecasting, mixed precision\gh{2883} support for faster training and prediction; hyperparameter search for fine-tuning; model ensembling for improved accuracy and robustness; conformal prediciton for calibrated prediction intervals; and more.
We encourage readers to check out the \href{https://unit8co.github.io/darts/examples/25-FoundationModel-examples.html}{\book~documentation} for details and examples.

\section{Usage Example}
The following snippet shows an example of TSFM usage in \Darts, where Chronos-2 is fine-tuned on Australian beer production data~\cite{hyndman2021forecasting} using the month attribute as a future covariate.
The model is probabilistic and forecasts quantiles [10\%, 50\%, 90\%].

\begin{figure}[t]
\begin{quickstart}[Usage Example]
\begin{small}
\begin{minted}{python}
from darts.datasets import AusBeerDataset
from darts.models import Chronos2Model
from darts.utils.likelihood_models import (
    QuantileRegression as QR
)
from darts.utils.timeseries_generation import (
    datetime_attribute_timeseries as dat
)

# Load and split the data
series = AusBeerDataset().load().astype("f")
train, test = series[:-12], series[-12:]

# Create future covariates -- month of the year
future_covs = dat(series, "month", dtype="f")

# Define the fine-tunable probabilistic model
model = Chronos2Model(
    input_chunk_length=24,
    output_chunk_length=12,
    likelihood=QR([0.1, 0.5, 0.9]),
    enable_finetuning={"unfreeze": ["output_*"]},
    n_epochs=10
)

# Fine-tune the model
model.fit(train, future_covariates=future_covs)

# Forecast Mode 1: Direct Parameter Prediction
forecast = model.predict(
    n=12, predict_likelihood_parameters=True
)
# Forecast Mode 2: Monte Carlo Sampling
forecast = model.predict(n=12, num_samples=100)
\end{minted}
\end{small}
\end{quickstart}
\vspace{-10mm}
\end{figure}

\section{Validation}
We validated the \Darts~implementations of TSFMs on \texttt{fev-bench-mini}~\cite{shchur2025fev}, which contains 20 out of 100 datasets from the full \texttt{fev-bench} for quick, reproducible evaluation.
PatchTST-FM is omitted as it did not participate in the original evaluation.
Table~\ref{tab:validation} reports the skill scores of the \Darts~implementations and the original implementations.
Despite some minor differences (some models can use longer contexts without needing to call \texttt{fit}), the results are broadly consistent, with deviations of at most 1.3\%.
Reproduction code and full results are available at \href{https://github.com/dennisbader/darts-foundation-model-paper}{\github~dennisbader/darts-foundation-model-paper}.

\begin{table}[H]
  \caption{Validation results on \texttt{fev-bench-mini}. Scaled Quantile Loss (SQL) for probabilistic forecasts and Mean Absolute Scaled Error (MASE) for point forecasts are reported as skill scores, i.e., error reduction relative to a seasonal naive baseline. Higher is better, and 0\% indicates parity with the baseline.}
  \label{tab:validation}
  \begin{center}
    \begin{small}
\begin{tabular}{@{}ll|rrrr@{}}
\toprule
\multirow{2}{*}{Library} & \multirow{2}{*}{Model} & \multicolumn{2}{c}{Skill Score (\%)} & \multicolumn{1}{c}{\multirow{2}{*}{\begin{tabular}[c]{@{}c@{}}Leaked\\ Datasets\end{tabular}}} & \multicolumn{1}{c}{\multirow{2}{*}{Fails}} \\ \cmidrule(lr){3-4}
 &  & \multicolumn{1}{c}{SQL} & \multicolumn{1}{c}{MASE} & \multicolumn{1}{c}{} & \multicolumn{1}{c}{} \\ \midrule
\multirow{3}{*}{\Darts} & Chronos-2 & 50.4 & 39.3 & 0 & 0 \\
 & TiRex & 43.2 & 31.1 & 0 & 0 \\
 & TimesFM 2.5 & 42.8 & 31.0 & 1 & 0 \\ \midrule
\multirow{4}{*}{Original} & Chronos-2 & 51.1 & 40.4 & 0 & 0 \\
 & TiRex & 43.4 & 31.4 & 0 & 0 \\
 & TimesFM 2.5 & 44.0 & 32.3 & 1 & 0 \\
 \bottomrule
\end{tabular}
    \end{small}
  \end{center}
\vspace{-10mm}
\end{table}

\section{Conclusion}
\Darts~now offers a unified foundation for harnessing time series foundation models, conforming to established API standards while adding new capabilities for fine-tuning and explainability.
Current limitations include the small number of supported TSFMs (four at launch) and the absence of parameter-efficient fine-tuning methods such as LoRA~\cite{hu2022lora}.
We continue to expand \Darts~with more features and models, and welcome contributions from all users as well as TSFM providers.
If you have not used the framework before, we encourage you to try it out, share your feedback on \href{https://github.com/unit8co/darts/issues}{\github~GitHub}, and contribute if able.

\section*{Acknowledgements}
We would like to acknowledge Lukas Fischer and Martin Loretz for contributing the TiRex~\cite{auer2025tirex} implementation~\gh{3038} in \Darts;
Abdul Fatir Ansari for the review and feedback on the Chronos-2~\cite{ansari2025chronos2} implementation and \texttt{FoundationModel} design~\gh{2944};
Rajat Sen and Dustin Brunner for guiding the TimesFM-2.5~\cite{das2024timesfm} implementation~\gh{2980}.
We also sincerely thank the \Darts~community for the feedback and contributions to the library, which have been instrumental in its growth and improvement.

\bibliography{paper}
\bibliographystyle{icml2026}

\end{document}